\documentclass{article}

\usepackage{microtype}
\usepackage{graphicx}
\usepackage{subfigure}
\usepackage{booktabs} 

\usepackage{hyperref}

\usepackage[accepted]{icml2021}
\usepackage{wrapfig}
\usepackage{amsmath}
\usepackage{algorithmic} 
\usepackage{algorithm} 
\usepackage{soul}

\usepackage{paralist, tabularx}
\definecolor{red}{RGB}{255, 0, 0}
\definecolor{blue}{RGB}{0, 0,255}

\definecolor{orange}{RGB}{179, 98, 0}

\icmltitlerunning{Batch Curation for Unsupervised Contrastive Representation Learning}

\begin{document}

\twocolumn[
\icmltitle{Batch Curation for Unsupervised Contrastive Representation Learning}



\icmlsetsymbol{equal}{*}

\begin{icmlauthorlist}
\icmlauthor{Michael C. Welle}{equal,kth}
\icmlauthor{Petra Poklukar}{equal,kth}
\icmlauthor{Danica Kragic}{kth}
\end{icmlauthorlist}

\icmlaffiliation{kth}{RPL, KTH Royal Institute of Technology, Sweden}
\icmlcorrespondingauthor{Michael C. Welle}{mwelle@kth.se}

\icmlkeywords{Machine Learning, ssl workshop, batch curation, unsupervised representation learning}

\vskip 0.3in
]



\printAffiliationsAndNotice{\icmlEqualContribution} 

\begin{abstract}
The state-of-the-art unsupervised contrastive visual representation learning methods that have emerged recently (SimCLR, MoCo, SwAV) all make use of data augmentations in order to construct a pretext task of instant discrimination consisting of similar and dissimilar pairs of images.
Similar pairs are constructed by randomly extracting patches from the same image and applying several other transformations such as color jittering or blurring, while transformed patches from different image instances in a given batch are regarded as dissimilar pairs. We argue that this approach can result 
similar pairs that are \textit{semantically} dissimilar.
In this work, we address this problem by introducing a \textit{batch curation} scheme that selects batches during the training process that are more inline with the underlying contrastive objective. We provide insights into what constitutes beneficial similar and dissimilar pairs
as well as validate \textit{batch curation} on CIFAR10 by integrating it in the SimCLR model.

\end{abstract}

\section{Introduction}
\label{intro}
Extracting data representation that are compact, expressive, and meaningful is a long term goal in the field of machine learning~\cite{bengio2013representation}. In recent years, most of the efforts have been put on learning general data representations that are suitable for a variety of different downstream task without using human annotation. 
In computer vision, prior work addressed this by solving pretext tasks defined in a self-supervised manner such as denoising autoencoders~\cite{vincent2008extracting}, solving jigsaw puzzles~\cite{noroozi2016unsupervised}, or motion segmentation~\cite{pathak2017learning} 
. A specific pretext task, namely instance discrimination~\cite{wu2018unsupervised}, has recently emerged as the state-of-the-art for unsupervised visual representation learning, spawning a polyhedral of new models based on contrastive learning such as SimCLR~\cite{chen2020simple}, MoCo~\cite{he2020momentum} or SwAV~\cite{caron2020unsupervised}.

In these methods, the pretext tasks are constructed by applying various data transformations, for example, cropping, color jittering or gaussian blur, in order to obtain similar and dissimilar pairs in an unsupervised way. Thanks to the thorough ablation study performed by the respective methods, it is established that \textit{random resized cropping}, i.e., extracting a patch of random size from the original image and resizing it back to the size of the original image, is one of the crucial data augmentation for learning useful representations~\cite{chen2020simple}. The intuition behind it is that patches drawn from the same image contain similar semantic information and should therefore be encoded close to each other. 
However, we hypothesise that this approach 
can produce patches containing semantically dissimilar information which decelerate learning as they violate this core similarity assumption.



\begin{figure*}[t]
\centering
    \includegraphics[width=0.9\linewidth]{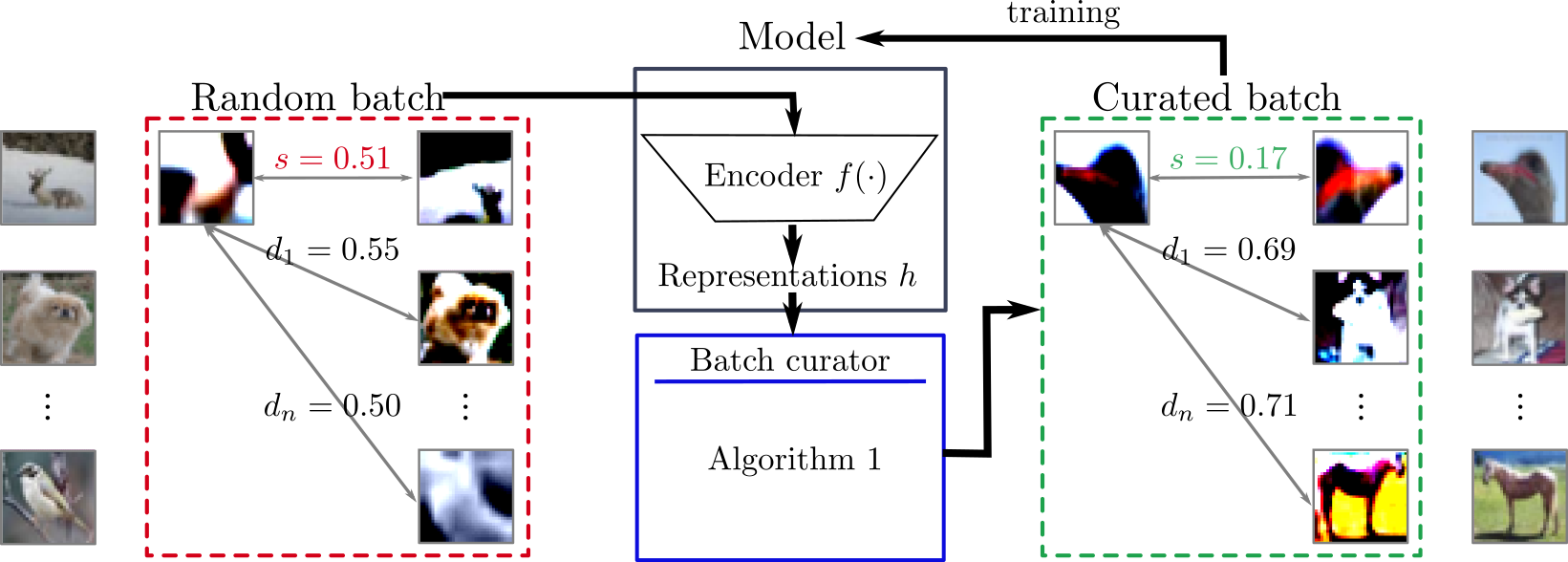}
    \caption{Overview of the proposed batch curation scheme. The curator (middle, blue) uses the model in training (middle, black) to curate a randomly sampled batch (left, red) of images such that the distance between the similar pairs $s$ is smaller than the distance $d_i$ among any dissimilar pair. The curated batch (right, green) is then used for training of the model. The original images from which patches in the batches are extracted are shown on the left and right. To obtain meaningful approximation of the distances among similar and dissimilar pairs, the
    curation scheme is activated after a \textit{warm-up} period. See Section~\ref{sec:curation} for more information.} 
    \label{fig:overview}
\end{figure*}

In this work, we first take a closer look at the random resize cropping and investigate how the size of the patches as well as their relative position to each other in the original image influences the learning process. Based on our findings, we then propose a \textit{batch curation} scheme, which during training constructs similar and dissimilar pairs that are aligned with the contrastive objective. Our scheme is simple as it uses the given model in training to self-curate batches such that similar patches are semantically close as judged by the model itself.
It can be easily adapted to any contrastive learning setting with arbitrary number of similar and dissimilar pairs, and can therefore be integrated into any of the current state-of-the-art methods. We validate our batch curation scheme in a small scale experiment by integrating it into the SimCLR~\cite{chen2020simple} model trained using CIFAR-10~\cite{krizhevsky2009learning} dataset. We show an improvement of $\approx 1.5\%$ on the K-NN classification evaluation as well as competitive performance on the linear classification protocol. To the best of our knowledge, we are the first to introduce a procedure for providing training batches of higher quality for unsupervised visual contrastive learning methods.

\section{Investigation of Patch Configurations}
In this section, we perform a deeper investigation on how the patch size and their pairwise relative position in the original image influences the learning process. Our analysis is based on extracting two patches from the same image similar to the approach taken in~\cite{chen2020simple}. All the investigations in this paper are performed on the CIFAR-10~\cite{krizhevsky2009learning} dataset containing images of dimension $32 \times 32$.


\subsection{Patch Configurations}
\begin{figure}[h]
    \includegraphics[width=\linewidth]{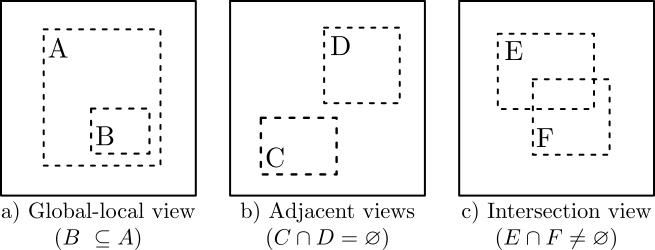}
    \caption{Three different configurations of two patches drawn from an image. From left to right: a) global-local view, b) adjacent view, and c) intersection view. } 
    \label{fig:patches_config}
\end{figure}

We first take a closer look into possible relative positions that two patches extracted from an image can attain. In Fig.~\ref{fig:patches_config} we show the three possible configurations:  \emph{a)} \textit{global-local view} where one patch is completely contained in the other,  \emph{b)} \textit{adjacent view} where patches have no intersection, and \emph{c)} \textit{intersection view} where patches have a non-empty intersection but are not contained in one another.

\begin{figure}[h]
    \includegraphics[width=\linewidth]{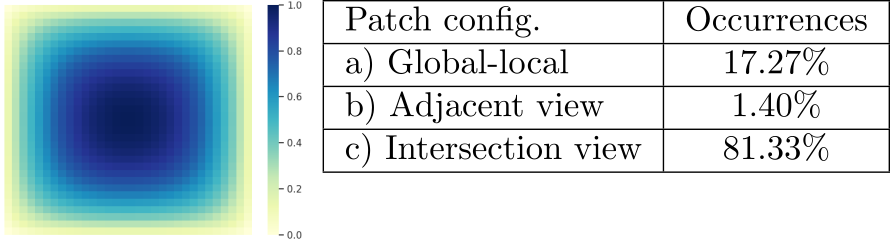}
    \caption{Normalised coverage heatmap of one million randomly drawn patches (left) and their patch configurations (right).} 
    \label{fig:patches}
\end{figure}

We perform random resized cropping using default scale $[0.08,1]$ and range $[0.75, 1.33]$ parameters in Pytorch implementation\footnote{\url{https://pytorch.org/vision/stable/transforms.html\#torchvision.transforms.RandomResizedCrop}} one million times on CIFAR-10 images, and show the occurrence of the three different configurations in the right of Fig.~\ref{fig:patches}. We can see that the majority of patch configurations are intersection views ($81.33\%$), while the global-local views occur in $17.27\%$ of cases. The adjacent views occur in only $1.4\%$ of cases which is expected since the average patch size is  $49\%$ of the original image.
On the left of Fig.~\ref{fig:patches} we display a normalised coverage heatmap of the sampled patches onto an image where pixels in blue denote the ones that are covered the most. As expected, we can observe a clear bias towards the middle of the image, which  
might be more beneficial for object centered datasets such as CIFAR-10 and ImageNet~\cite{russakovsky2015imagenet}.

\subsection{Performance of Different Patch Configurations on SimCLR} \label{sec:patch_conf_simclr}

Given the insights of patch configuration occurrences, we investigate the performance of a SimCLR when limiting the patches we sample to either \emph{i)} only global-local views, \emph{ii)} only adjacent views, \emph{iii)} only intersection views, and \emph{iv)} a version where all configurations occur with equal probability. We trained all models for $500$ epochs using batch size $128$, feature dimension $128$, and temperature $0.5$. We evaluated them using K-NN evaluation where $K = 200$, and linear evaluation protocol where one linear layer is trained on the obtained representations for $100$ epochs. 

We show in Table~\ref{tab:patch_results} that the best performing models are the one using only intersection views and the default SimCLR (both obtaining $\approx 90.5\%$ on linear evaluation and $86.5\%$ on K-NN evaluation). As we showed in Fig.~\ref{fig:patches}, the latter naturally contains a high percentage of intersection views. Enforcing just global-local patches as well as using an equal amount of each configuration performs slightly worse, while the model employing only adjacent views performs the worst (only $\approx 65\%$ on both linear and K-NN evaluations). We observed that both the model containing equal amount of each configuration and using adjacent views have smaller average patch size.

\begin{table}[h]
\resizebox{\linewidth}{!}{
\begin{tabular}{|l|c|c|c|}
\hline
SimCLR model version                        & Linear acc. & K-NN acc. & avg. p. size \\ \hline
\multicolumn{4}{|c|}{Different patch configurations}           \\ \hline
\textbf{default}             &   $\boldsymbol{90.67 \%}$    &   $\boldsymbol{86.29\%}$    &        $\boldsymbol{49\%}$         \\ \hline
global-local        &   $87.32\%$    &  $81.95 \%$     &     $51\%$            \\ \hline
adjacent view       &  $64.59\%$     &  $65.96 \%$     &     $17\%$            \\ \hline
\textbf{intersection view}   &    $\boldsymbol{90.72\%}$   &  $\boldsymbol{86.63 \%}$     &     $\boldsymbol{49\%}$            \\ \hline
equal configuration      &   $88.60\%$     &  $83.95 \%$    &        $39\%$         \\ \hline
\multicolumn{4}{|c|}{Different patch sizes}                    \\ \hline
big patches         &  $89.17\%$    &   $85.99\%$     &       $70\%$          \\ \hline
small patches         &  $23.24\%$    &   $76.31\%$     &       $29\%$          \\ \hline
global-local big patches &  $86.50\%$     &  $82.87\%$     &      $73\%$           \\ \hline
\end{tabular}
}
\caption{Best accuracy of SimCLR models trained with different patch configurations on a test set using K-NN evaluation (left) as well as linear evaluation (middle). Right column shows the average patch size in relation to the original image size.}
\label{tab:patch_results}
\end{table}

\begin{figure}[t]
  \begin{center}
    \includegraphics[width=0.98\linewidth]{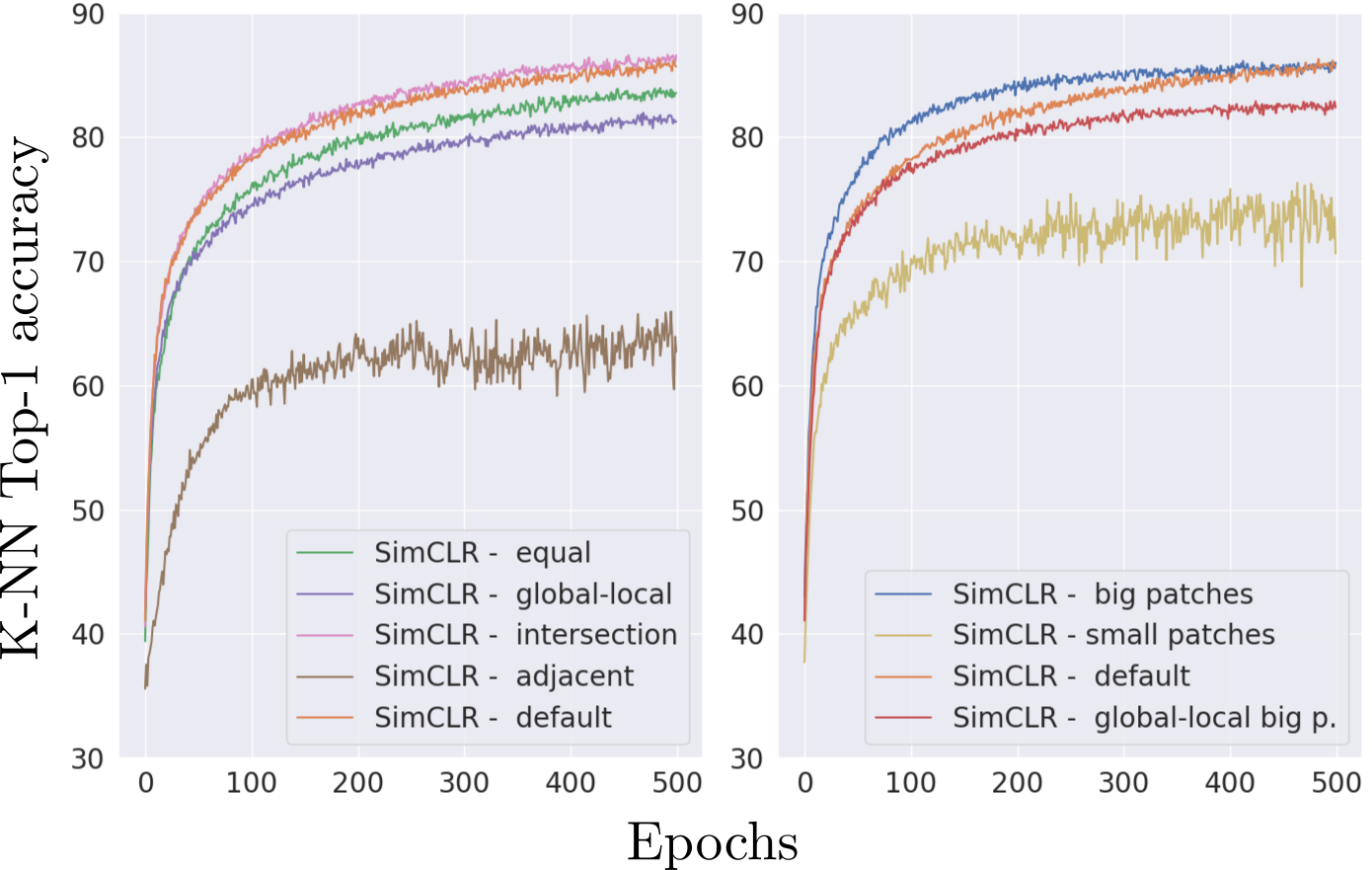}
    \caption{Top-1 accuracy of SimCLR models obtained during training with different patch configurations on a test set using K-NN evaluation. } 
    \label{fig:patches_perform}
  \end{center}
\end{figure}

In order to further investigate the effect of different patch sizes, we evaluate three additional models: (i) the default SimCLR, (ii) a global-local version both with scale $[0.5,1]$ producing larger patches, and (iii) the default SimCLR with scale $[0.08,0.5]$ producing smaller patches.
Fig.~\ref{fig:patches_perform} shows the K-NN Top-1 accuracy obtained during training for all models. We observe that bigger patches lead to a faster increase in performance but do not yield a better performance in the long run. We also see that the big patch version of global-local views performs slightly better than the default global-local in the K-NN evaluation but slightly worse in the linear evaluation protocol (see Table~\ref{tab:patch_results}). The default SimCLR with small patches performs significantly worse obtaining only $\approx 23\%$ accuracy on the linear evaluation. In summary, we find that:
\begin{compactitem}
    \item Adjacent views seem to negatively affect the resulting representations possibly due to their small size. 
    \item Larger patches on average give a boost at the beginning of the training but do not increase the overall performance.
    \item Intersecting views are more beneficial than global-local or adjacent views for the SimCLR model.
\end{compactitem}

\section{Batch Curation} \label{sec:curation}

Motivated by the  differences in the performance of the SimCLR models when varying the type of patch configuration presented in the previous section, we hypothesise more generally that the quality of batches used during training of any unsupervised contrastive  visual representation learning model is directly correlated to the quality of the learned representations.

More precisely, from the poor performance of the adjacent configuration of patches in Table~\ref{tab:patch_results}, we hypothesise that sampling small patches that do not contain any relevant semantic information hinders the learning process of the model in training.
We conjecture that this results in an ill-defined pretext task where patches that are labeled similar are in fact not semantically similar. 
To this end, we propose  a \textit{batch curation} scheme presented in the next section.

\subsection{The Batch Curator}

The proposed \textit{batch curation scheme} curates a randomly sampled batch such that similar pairs of patches are always closer together than any of the dissimilar pairs. We visualize its principles in Fig.~\ref{fig:overview} using a contrastive learning setting with one similar pair and several dissimilar pairs\footnote{Note that this can be easily adapted to arbitrary number of similar or dissimilar pairs.}.

The details of our procedure are outlined in Algorithm~\ref{alg::curator}. Firstly, we encode the randomly sampled patches using the model trained in the previous epoch to obtain their representations. 
Secondly, we use these representations to calculate the distances $M_\mathcal{S}, M_\mathcal{D}$ between similar and dissimilar pairs, respectively. 
We accept the batch if the largest distance between similar pairs $d_\mathcal{S}$ is smaller that the smallest distance among the dissimilar pairs $d_\mathcal{D}$. If this criteria is not fulfilled, we re-sample pairs that violate it until we obtain an adequate batch. The curated batch is then used to perform one step of the gradient-decent in the training of the unsupervised contrastive learning model. In order to obtain somewhat meaningful distances in the learned representation space, we start the batch curation scheme after an initial warm-up period of $w$ epochs. Since the curation is simply performed by the model itself, it can be easily integrated to any unsupervised contrastive representation learning method.


\begin{algorithm}[t]
\caption{General batch curation scheme.}
\begin{algorithmic}[1]
\REQUIRE Set of similar images $\mathcal{S}$
\REQUIRE Set of dissimilar images $\mathcal{D}$
\STATE $d_\mathcal{S} = d_\mathcal{D} \gets 0$
\WHILE {not $d_\mathcal{S} < d_\mathcal{D}$}
\STATE $h_\mathcal{S} \gets f(\mathcal{S}), h_\mathcal{D} \gets f(\mathcal{D})$  [extract representations]
\STATE $M_\mathcal{S} \gets \texttt{compute distance}(h_\mathcal{S}, h_\mathcal{S})$ [distance among similar pairs]
\STATE $M_\mathcal{D} \gets \texttt{compute distance}(h_\mathcal{S}, h_\mathcal{D})$ [distance among dissimilar pairs]
\STATE $d_\mathcal{S} \gets \max(M_\mathcal{S})$ 
\STATE $d_\mathcal{D} \gets \min(M_\mathcal{D})$ 

\STATE {resample images in $\mathcal{S}$ and $ \mathcal{D}$ violating $d_\mathcal{S} < d_\mathcal{D}$}
\ENDWHILE
\STATE return $\mathcal{S}, \mathcal{D}$
\end{algorithmic}
\label{alg::curator}
\end{algorithm}



\begin{table}[h]
\resizebox{\linewidth}{!}{
\begin{tabular}{|l|c|c|c|}
\hline
SimCLR model version                        & Linear acc. & K-NN acc. & avg. p. size \\ \hline

default w/o batch curator &   $90.67 \%$    &   $86.29\%$    &        $49\%$         \\ \hline
\textbf{with batch curator}       &   \boldmath$90.81\%$    &  \boldmath$ 87.63\%$     &     $49\%$            \\ \hline   
\end{tabular}
}
\caption{Best accuracy of SimCLR models trained with batch curation scheme (bottom) and without it (top) on a test set using K-NN evaluation (left) as well as linear evaluation (right). Right column shows the average patch size in relation to the image size.}
\label{tab:patch_cur_results}
\end{table}

\subsection{Performance of the Batch Curator on SimCLR}

\begin{wrapfigure}{rt}{0.58\linewidth}
  \begin{center}
    \includegraphics[width=0.98\linewidth]{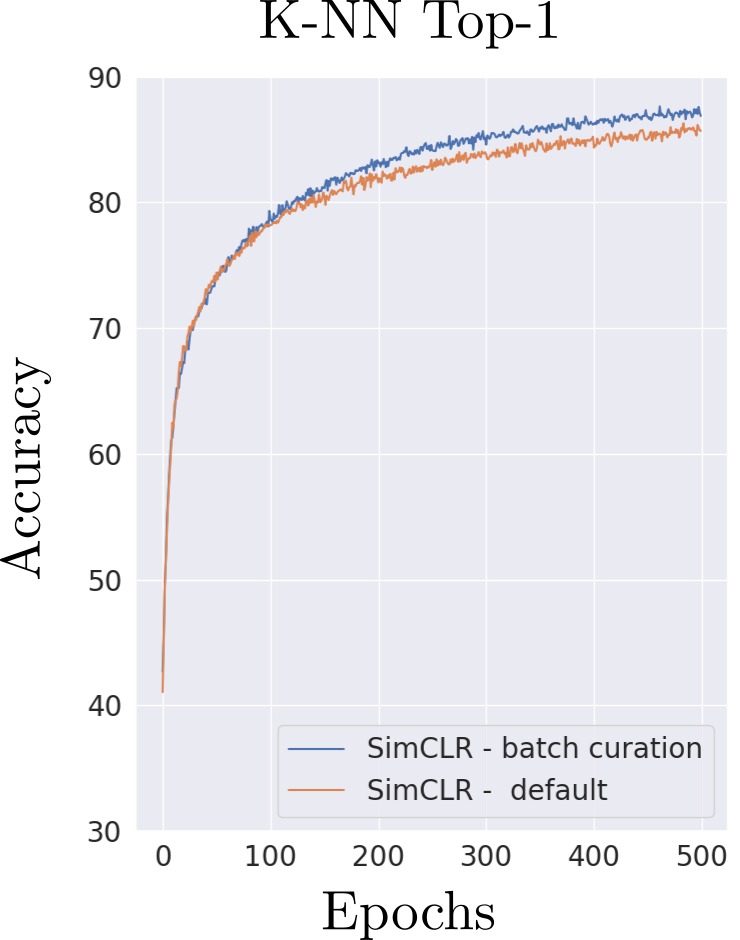}
    \caption{Top-1 accuracy of SimCLR model trained with batch curation scheme (blue) and without it (orange) using K-NN evaluation. } 
    \label{fig:patches_cur_perform}
  \end{center}
\end{wrapfigure}

We validate our approach by comparing the performance of a SimCLR model trained with the proposed batch curation scheme (SimCLR-batch curation) and one without it (SimCLR-default). We trained and evaluated both models as described in Section~\ref{sec:patch_conf_simclr}. 
The evaluation results are shown in Table~\ref{tab:patch_cur_results}, where we can observe a $1.5 \%$ improvement in the K-NN evaluation and a slight improvement in the linear evaluation. In Fig~\ref{fig:patches_cur_perform}, we show the changes in K-NN Top-1 accuracy of the test set during training. We can clearly see the improvement in the performance when batch curation is started, showing that it leads to improved representations.

We emphasise that the presented experimental results do not match the state-of-the-art $92\%$ accuracy on the linear evaluation on CIFAR-10 as reported in~\cite{chen2020simple} due to the limited hardware setup. Note that even when using batch size $128$, it takes more than $20$h to train each SimCLR model using a single GPU. In future, we plan to validate our batch curation scheme on more complex datasets such as ImageNet
as well as integrate it into other state-of-the-art unsupervised contrastive representation learning methods such as MoCo and SWaV. However, we leave the extensive scaling, for example as performed in~\cite{goyal2021self} using up to 1.3 billion model parameters and 512 GPUs, to other research groups with larger computational resources.

\section{Conclusion and Future Work}

In this work, we investigated the common transformation applied to images in the unsupervised contrastive  visual representation learning domain, namely random resize cropping. We showed that the most beneficial patch configuration consists of intersecting patches, while smaller patches seem to be detrimental. We hypothesise that the latter occurs because patches do not contain enough semantic information. To this end, we proposed a simple batch curator scheme, which curates the training batches to be more aligned with the contrastive learning objective. We show that our scheme improves the end performance when integrated to the SimCLR model on CIFAR-10 dataset. As future work, we plan to perform extensive fine-tuning of our scheme, scaling it to larger datasets such as ImageNet and integrating it to other existing methods.





\bibliography{main}
\bibliographystyle{icml2021}


\end{document}